\documentclass[]{article}
\usepackage{proceed2e}
\usepackage[breaklinks=true]{hyperref}
\usepackage{url}
\usepackage{breakurl}
\usepackage{balance}
\usepackage{times}
\usepackage{graphicx} 
\usepackage{subfigure} 
\usepackage{amsmath}
\usepackage{amssymb}
\usepackage{auto-pst-pdf}
\usepackage{pst-node,pstricks}
\usepackage{wrapfig}

\newcommand{\bi}{{\bf i}}
\newcommand{\bj}{{\bf j}}
\newcommand{\bk}{{\bf k}}
\newcommand{\bl}{{\bf l}}

\newcommand{\bw}{{\bf w}}
\newcommand{\bE}{{\bf E}}
\newcommand{\bW}{{\bf W}}
\newcommand{\cut}[1]{}

\usepackage{algorithmic}

\title{Hierarchical learning of grids of microtopics}

\author{ {\bf Nebojsa Jojic\thanks{Alessandro and Nebojsa equally contributed to this work}} \\
Microsoft Research, \\
Redmond, WA 98052 \\
\And
{\bf Alessandro Perina$^*$}  \\
WDG Core Data Science, Microsoft \\
Redmond, WA 98052 \\
\And
{\bf Dongwoo Kim}   \\
Australia National University\\
Canberra, Australia
}

\begin{document}

\maketitle

\begin{abstract}
The counting grid is a grid of \emph{microtopics}, sparse word/feature distributions. The generative model associated with the grid does not use these microtopics individually, but in predefined groups which can only be (ad)mixed as such. Each allowed group corresponds to one of all possible overlapping rectangular windows into the grid. The capacity of the model is controlled by the ratio of the grid size and the window size.
This paper builds upon the basic counting grid model and it shows that hierarchical reasoning helps avoid bad local minima, produces better classification accuracy and, most interestingly, allows for extraction of large numbers of coherent microtopics even from small datasets. We evaluate this in terms of consistency, diversity and clarity of the indexed content, as well as in a user study on word intrusion tasks. We demonstrate that these models work well as a technique for embedding raw images and discuss interesting parallels between hierarchical CG models and other deep architectures.
\end{abstract}
\section{INTRODUCTION}
\begin{figure*}[t!]
\centering
\includegraphics[width=1\textwidth]{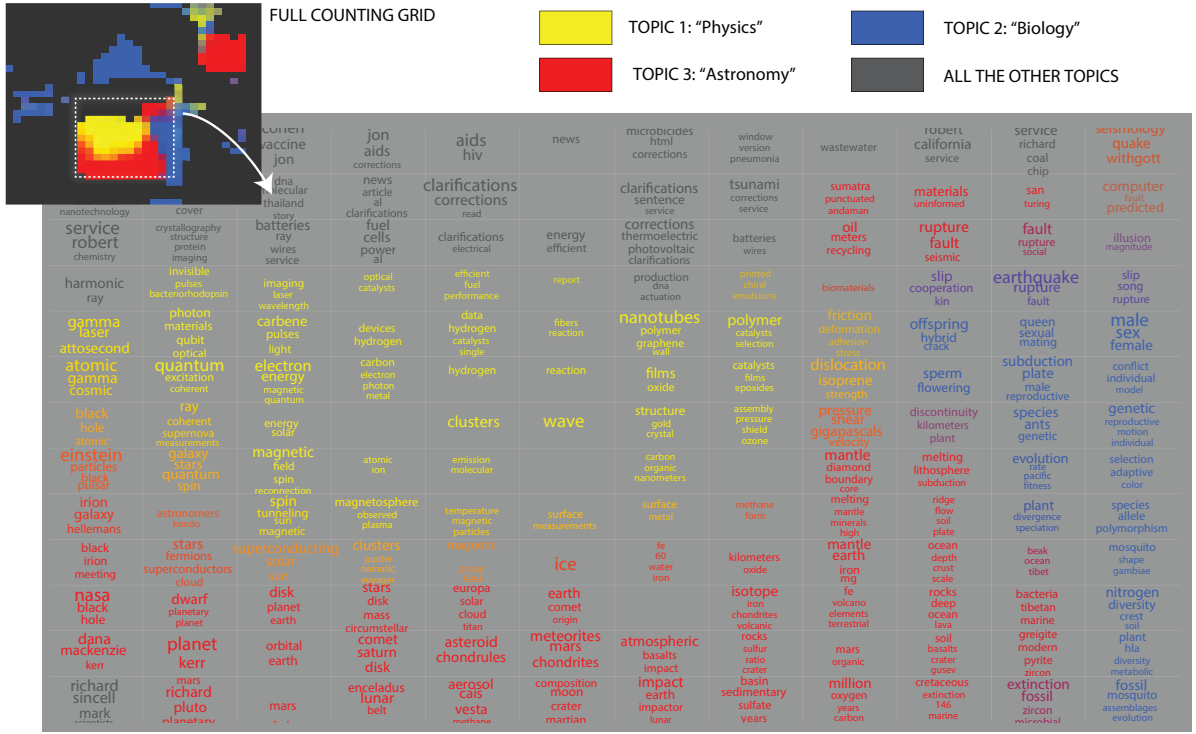}
\vspace{-0.6cm}
\caption{Clash of topics: LDA topics are mapped onto  a counting grid. As shown in the top left panel, LDA's topics cluster in contiguous areas on the grid. In the enlarged part of the grid, for each microtopic we show the most likely words if they exceed a threshold.}
\vspace{-0.4cm}
\label{fig:fig0}
\end{figure*}

\cut{Machine learning has recently entered a renaissance both in terms of research breakthroughs and wide spread of practical uses. Both research and applications, however, mostly focus on supervised settings \cite{deeplearning_nature}. Unsupervised learning remains the holy grail of machine learning research, or at least a much needed step toward strong AI. In practice, unsupervised learning is equally desirable, as the massive and ever growing amount of data created through internet activity and increasing diversity of physical sensors obviously comes unlabeled. In fact labels for some aspects of new data often have not even been invented yet. In particular, unstructured text that is created online in forums, social media, product reviews, user feedback, blogs, etc. is of great interest. These sort of data need to be processed by unsupervised learning algorithms that are preferably well suited to user interface strategies so that humans are aided in a daunting task of filtering and comprehending this vast amount of information.}
Recently, a new breed of topic models, dubbed counting grids (CG) \cite{cgUai,ccg}, has been shown to have advantages in unsupervised learning over previous topic models, while at the same time providing a natural representation for visualization and user interface design \cite{perinaKDD}. CG models are \emph{generative} models based on a grid of word distributions, which can best be thought of as the grounds for a massive Venn diagram of documents. The intersections among multiple documents (bags of words) create little intersection units with a very small number of words in them (or rather, a very sparse distribution of the words). The grid arrangement of these sparse distributions, which we will refer to here as \emph{microtopics}, facilitates fast cumulative sum based inference and learning algorithms that chop up the documents into much smaller constitutive pieces than what traditional topic models typically do. For example, Fig. \ref{fig:fig0} shows a small part of such a grid with a few representative words with greatest probability from each microtopic. Each of the Science magazine abstracts used to train this grid is assumed to have been generated from a group of microtopics found in a single 4 $\times$ 4 window with equal weight given to all component microtopics. Thus, each microtopic can be 16 times sparser than the set of documents grouped into the window. 

A document may share a window with another very similar document, but it is also mapped so that it only partially overlaps with a window that is the source for a set of slightly less related documents. The varying window overlap literally results in a varying overlap in document themes. This modeling assumption results in a trained grid where nearby microtopics tend to be related to each other as they are often used together to generate a document. Consider, e.g., the lower right 4$\times$4 window in Fig. \ref{fig:fig0}. The word distributions in these 16 cells are such that a variety of Science papers on evidence of ancient life on Earth could be generated by sampling words from there. (Note that each cell, though of very low entropy, contains a distribution over the entire vocabulary.) In the posterior distribution, this window is by far the most likely source for an article on a bizarre microorganism that produced nitrogen in cretaceous oceans. In the 4$\times$4 window two cells to the left of this example we find mapped a variety of articles on even more ancient events on Earth, e.g. on how sulfur isotopes reveal a deep mantle storage of ancient crust. But there we also start to see words which increase the fit for articles that describe similar events on other planets. Further movement to the left gets us away from the Earth and into astronomy. 

To demonstrate the refinement of the microtopics compared to topics from a typical topic model, the color labeling of the grid was created so as to reflect the Kullback–-Leibler (KL) divergence of the individual microtopics to the topics trained on the same data through latent Dirichlet allocation (LDA). The LDA topics, hand-labeled after unsupervised training, correspond to fairly broad topics, while the CG represents the data as a group of slowly evolving microtopics. For example, all the yellow coded microtopics map to the "Physics" LDA topic, but they occupy a contiguous area in which from left to right the focus slowly shifts from electromagnetism and particle physics to material science. Furthermore, it is interesting to see the microtopics that occupy the boundaries between coarser topics that LDA model found, capturing the links among astronomy, physics and biology. It is immediately evident that the 2D CGs can have great use in data visualization, though the model can be trained for arbitrary dimensionality \cite{cgUai}. These models combine topic modeling and data embedding ideas in a way that facilitates intuitive regularization controls and allows creation of much larger sets of organized sparse topics. Furthermore, they lend them selves to elegant visualization and browsing strategies, and we encourage the reader to see the example \url{http://research.microsoft.com/en-us/um/people/jojic/CGbrowser.zip}.

However, the existing EM algorithm for CG learning is prone to local minima problems which occasionally lead to under performance \cite{kusner-etal,ssvi}. In addition, no direct testing of the microtopic coherence has been performed to date, which makes it unclear if they are meaningful outside their windowed grouping. After all, a variety of sophisticated topic models have been developed and tested by the research community, but LDA seems to still beat them often in practice. E.g., [16,17] raise doubts that various reported perplexity improvements over the basic LDA model are meaningful as they are sensitive to smoothing constants in the model, and also fail to translate to improvements in human judgement of topic quality. In fact, LDA usually outperforms more complex models on tasks that involve human judgement, which may be the main reason why practitioners of data science prefer this basic model to others \cite{topic_models_practice}. Here we develop hierarchical versions of CG models, which in our experiments produced embeddings of considerably higher quality. We show that layering into deeper architectures primarily aids in avoiding bad local minima, rather than increasing representational capacity: The trained hierarchical model can be collapsed into an original counting grid form but with a much higher likelihood compared to the grids fit to the same data using EM with random restarts. The better data fit then translates into quantitatively better summaries of the data, as shown in numerical experiments as well as human evaluations of microtopics obtained through crowdsourcing.

\vspace{-0.3cm}
\section{HIERARCHICAL LEARNING OF GRIDS OF MICROTOPICS}
\vspace{-0.2cm}
\begin{figure}
\centering
\includegraphics[width = 1\columnwidth]{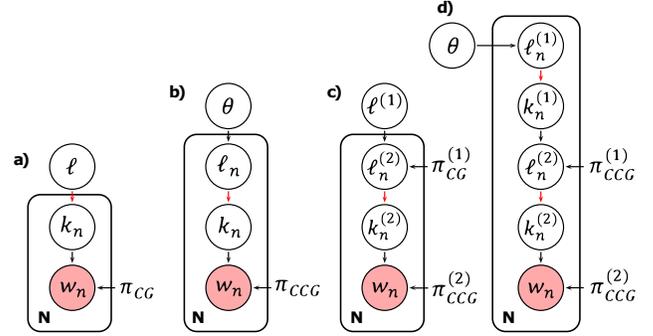}
\caption{\textbf{a)} The basic counting grid,  \textbf{b)} the componential counting grid, \textbf{c)} the hierarchical counting grid model (HCG) obtained by stacking a componential counting grid and a counting grid, and \textbf{d)} the hierarchical componential counting grid model (HCCG). Dotted circles represent the parameters of the models. Red links represents known conditional distributions $P(k_n|\ell_n) = U_{\ell}^W$ - Eq. \ref{eq:u}. They are distributions over the grid locations, uniformly equal to 1/$|\bW|$ in the window of size $\bW_\ell$ unequivocally identified by $\ell$.  }
\label{fig:gm}
\vspace{-0.4cm}
\end{figure}

\paragraph{The (C)CG grids \cite{cgUai,ccg}:} The basic counting grid $\pi_{\bk}$ \cite{cgUai} is a set of distributions on the $d$-dimensional toroidal discrete grid $\bE$ indexed by $\bk$. The grids in this paper are bi-dimensional and typically from $(E_x = 32) \times (E_y =32)$ to $(E_x = 64) \times (E_y = 64)$ in size. The index $z$ indexes a particular word in the vocabulary $z=[1\dots Z]$. Thus, $\pi_{\bi}(z)$ is the probability of the word $z$ at the $d$-dimensional discrete location $\bi$, and $\sum_z \pi_{\bi}(z)=1$ at every location on the grid. The model generates bags of words, each represented by a list of words $\bw = \{ w_n \}_{n=1}^N$ with each word $w_n$ taking an integer value between $1$ and $Z$.
The modeling assumption in the basic CG model is that each bag is generated from the distributions in a single window $\bW$ of a preset size, e.g., $W_x = 5\times W_y=5$. A bag can be generated by first picking a window at a $d$-dimensional location $\ell$, denoted as $W_{\ell}$, then generating each of the $N$ words by sampling a location $\bk_n$  for a particular microtopic $\pi_{\bk_n}$ uniformly within the window, and finally by sampling from that microtopic.
Because the conditional distribution $p(\bk_n|\ell)$ is a preset uniform distribution over the grid locations inside the window placed at location $\ell$, the variable $\bk_n$ can be summed out\cite{cgUai}, and the generation can directly use the grouped histograms
{\small
\begin{equation} \label{eq:h}
h_{\ell}(z)=\frac{1}{|\bW|} \sum_{\bj \in W_{\ell}} \pi_{\bj}(z),
\end{equation}
}
where $|\bW|$ is the area of the window, e.g. 25 when 5$\times$5 windows are used. 
In other words, the position of the window $\ell$ in the grid is a latent variable given which we can write the probability of the bag as
{\small
\begin{equation} \label{eq:cg}
P( \bw | \ell )= \prod_{w_n \in \bw} h_{\ell}(w_n) =\prod_{w_n \in \bw} \big( \frac{1}{|\bW|} \cdot \sum_{\bj \in W_\ell} \pi_{\bj}(w_n) \big)
\end{equation}
}
As the grid is toroidal, a window can start at any position and there is as many $h$ distributions as there are $\pi$ distributions. The former will have a considerably higher entropy as they are averages of many $\pi$ distributions.
Although the basic CG model is essentially a simple mixture assuming the existence of a single source (one window) for all the features in one bag, it can have a very large number of (highly related) choices $h$ to choose from. Topic models \cite{lda,ctm}, on the other hand, are admixtures that capture word co-occurrence statistics by using a much smaller number of topics that can be more freely combined to explain a single document.  
Componential Counting Grids (CCG) \cite{ccg} combine these ideas, allowing multiple groups of broader topics $h$ to be mixed to explain a single document. The entropic $h$ distributions are still made of sparse microtopics $\pi$ in the same way as in CG so that the CCG model can have a much larger number of topics than an LDA model without overtraining. More precisely, each word $w_n$ can be generated from a different window, placed at location $\ell_n$, but the choice of the window follows the same prior distributions $\theta_\ell$ for all words. Within the window at location $\ell_n$  the  word comes from a particular grid location $k_n$, and from that grid distribution the word is assumed to have been generated. The probability of a bag is now
{\small
\begin{equation} \label{eq:ccg}
P(\bw | \pi ) =  \prod_{w_n \in \bw} \sum_{\ell\in \bE}  \Big( \theta_\ell \cdot \big( \frac{1}{|\bW|} \sum_{\bj \in W_\ell} \cdot \pi_{\bj}(w_n) \big) \Big) 
\end{equation}
}
In a well-fit CCG model, each data point has an inferred $\theta_{\ell}$ distribution that usually hits multiple places in the grid, while in a CG, each data point tends to have a rather peaky posterior location distribution because the model is a mixture.   
Both models can be learned efficiently using the EM algorithm because the inference of the hidden variables, as well as updates of $\pi$ and $h$ can be performed using summed area tables \cite{Crow}, and are thus considerably faster than most of the sophisticated sampling procedures used to train other topic models. 
An intriguing property of these models is that even on a $32 \times 32$ grid with $1024$ microtopics $\pi$ and just as many grouped topics $h$, there is no room for too many independent groups. With a window size $8\times 8$, for example, we can place only $16$ windows without overlap, and the remaining windows are overlapping the pieces of these 16. The ratio between grid and window size is referred to as the \emph{capacity} of the model, and the training set size necessary to avoid overtraining the model only needs to be 1-2 orders of magnitude above the capacity number. Thus a grid of 1024 microtopics may very well be trainable with thousands of data points, rather than 100s of thousands that traditional topic models usually require for that many topics. 
\begin{figure*}[t!]
\centering
\includegraphics[width=0.9\textwidth]{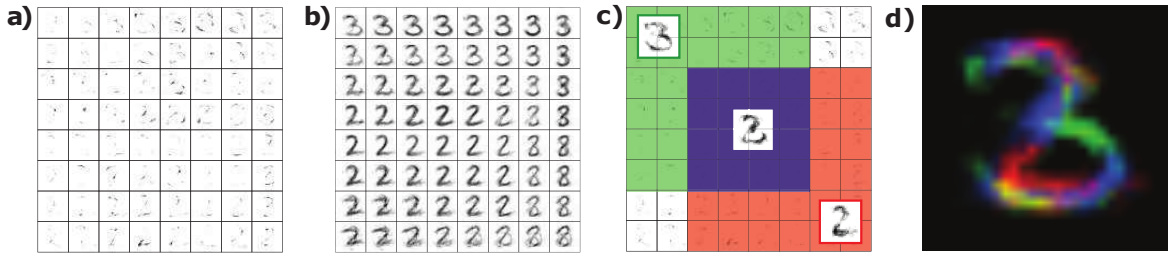}
\caption{Intersecting digits on a grid of strokes. Each digit image is represented by counts (intensity) associated with image locations. 
\textbf{a)} $\pi$-distributions \textbf{b)} $h$-distributions \textbf{c-d)} Intersecting digits}\vspace{-0.3cm}
\label{fig:dig}
\end{figure*}

\begin{figure*}[t!]
\centering
\includegraphics[width=1\textwidth]{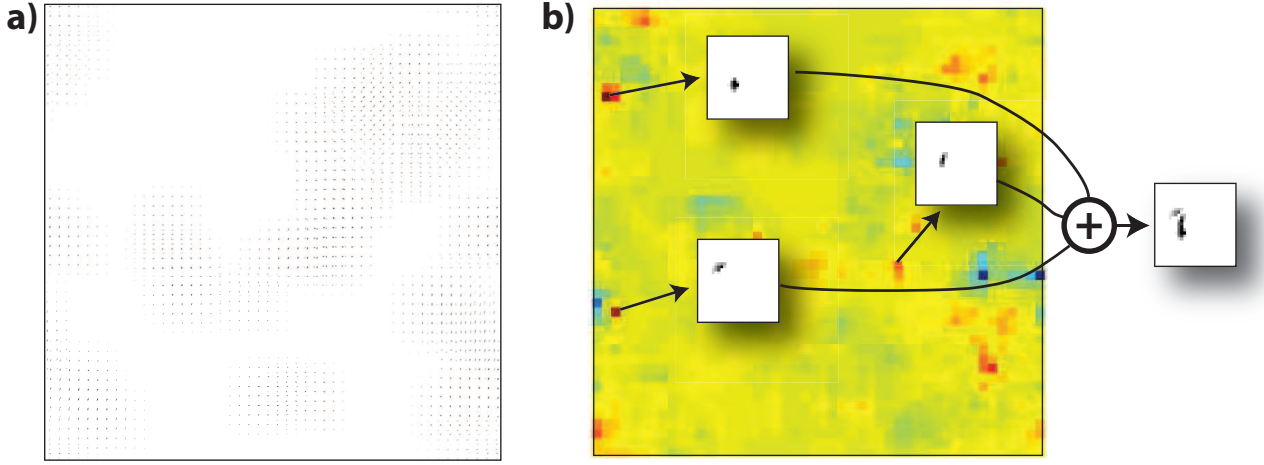}
\caption{The benefits of hierarchical learning: \textbf{a)} $h_{CCG}$ - a bigger higher resolution version in the appendix. \textbf{b)} Principal components of $\theta$ and three peaks put together. }
\label{fig:digits2}
\end{figure*}

\paragraph{Raw image embedding using (C)CGs:}
In previous applications of CG models to computer vision, images were represented as spatially disordered bags of features. We experimented with embedding raw images with full spatial information preserved, and we present this here as we feel that the image data helps in illuminating the benefits of hierarchical learning. An image described by a full intensity function $I(x,y)$ could be considered as a set of words, each word being an image location $z=(x,y)$. For a $N\times M$ image, we have a vocabulary of size $M\cdot N$. The number of repetitions of word $(x,y)$ is then set to be proportional to the intensity I(x,y). (In case of color images, the number of features is simply tripled with each color channel treated in this way). In other words, an unwrapped image is considered to be a word (location) histogram. $\pi$ and $h$ distributions can then also be seen as images, as they provide weights for different image locations. If we tile the image representations of these distributions we get additional insight into CGs as an embedding method.  
Fig. \ref{fig:dig} shows a portion of a $48\times48$ grid trained on 2000 MNIST digits assuming a $6\times 6$ window averaging. To illustrate the generative model, in c) we show the partial window sums for two overlapping windows over $\pi$. The green and blue areas form a window that generates a version of digit 3, which can be seen at the top left of this portion of the $h$ grid (panel b)). The blue and red, on the other hand, combine into a window that represent a digit 2 at the position (3,3) in panel b). Partial sums for green, blue and red areas are shown in c) and these partial sums, color coded and overlapped are also illustrated in d). Careful observation of b) or the full grid in the appendix, demonstrates the slow deformation of digits from one to another in the $h$ distributions. The appendix has additional examples of image dataset embedding, including rendered 3D head models and images of bold eagles retrieved by internet search. The CG $\pi$ distributions shown here look like little strokes, while $h$ distributions are full digits. The CCG model, on the other hand, combines multiple $h$ distributions to represent a single image, and so $h$ looks like a grid of strokes Fig. \ref{fig:digits2}a, while $\pi$ distributions are even sparser.

\paragraph{Hierarchical grids:} By learning a model in which microtopics join forces with their neighbors to explain the data, (C-)CG models tend to exhibit high degrees of relatedness of nearby topics. As we slowly move away from one microtopic, the meaning of the topics we go over gradually shifts to related narrowly defined topics as illustrated by Fig. \ref{fig:fig0}; this makes these grids attractive to HCI applications. But this also means that simple learning algorithms can be prone to local minima, as random initializations of the EM learning sometimes result in grouping certain related topics into large chunks, and sometime breaking these same chunks into multiple ones with more potential for suboptimal microtopics along boundaries.
To illustrate this, in Fig. \ref{fig:digits2}a we show a $48\times48$ grid of strokes $h$ (Eq. \ref{eq:h}) learned from 2000 MNIST digits using a CCG model assuming a $5\times 5$ window averaging. Nearby features $h$ are highly related to each other as they are the result of adding up features in overlapping windows over $\pi$ (which is not shown).
CCG is an admixture model, and so each digit indexed by $t$ has a relatively rich posterior distribution $\theta^t$ over the locations in the grid that point to different strokes $h$. In Fig. \ref{fig:digits2}, we show one of the main principal components of variation in $\theta$ as an image of the size of the grid. For three peaks there, we also show $h$-features at those locations. The combination of these three sparse features creates a longer contiguous stroke, which indicates that this longer stroke is often found in the data. Thus, the separation of these features across three distant parts of the map is likely a result of a local minimum in basic EM training. To transfer this reasoning to text models, consider the 5th cell in the first row in Fig. \ref{fig:fig0} with words HIV, AIDS, and the blue cell in the middle of the last column with words SELECTION, ADAPTIVE. The separation of these two things in faraway locations may very well be a result of a local minimum, which could be detected if location posteriors exhibit correlation. 
This illustration points to an idea on how to build better models. The distribution over locations $\ell$ that a data point $t$ maps to (a posteriori) could be considered a new representation of the data point (digit in this case), with the mapped grid locations considered as features, and the posterior probabilities for these locations considered as feature counts. Thus another layer of a generative model can be added to generate the locations in the grid below, Fig. \ref{fig:gm}c-d. It is particularly useful to use another microtopic grid model as this added layer, because of the inherent relatedness of the nearby locations in the grid. The layer above can thus be either another admixture grid model (Componential Counting Grid - CCG), or a mixture (CG), and this layering can be continued to create a deep model. As CG is a mixture model, it terminates the layering: Its posterior distributions are peaky and thus uncorrelated. However, an arbitrary number of CCGs can be stacked on top of each other in this manner, terminating on top with a CG layer to form a hierarchical CG (HCG) model, or terminating in a CCG layer to form a hierarchical CCG (HCCG) model. In each layer, the pointers to features below are grouped, which should result in creating a contiguous longer stroke as discussed above in a grid cell that contains a combination of pointers to the lower layers.

For the sake of brevity, we only derive the HCG learning algorithm with a single intermediate CCG layer. The extension to HCCG and higher order hierarchies is reported in  the appendix. Variational inference and learning procedure for counting grid-based models utilizes cumulative sums and is only slower than training an individual (C)CG layer by a factor proportional to the number of layers. 
The graphical model for HCG is shown in Fig. \ref{fig:gm}c, where location variables pointing to grids in different layers have the same name, $\ell$ but carry a disambiguating superscript. \emph{To avoid superscripts in the equations below, we renamed the CG's location variable from $\ell^{(1)}$ to $m$ and dropped the superscript ``$^{(2)}$' in the layer above}. The bottom CCG layer follows
{\small
\begin{equation}
P(w_n|k_n,\pi_{CCG}) = \pi_{CCG,k_n}(w_n)
\end{equation}
}
{\small
\begin{equation} \label{eq:u}
P(k_n |\ell_n ) = U_{\ell_n}^W(k_n) = \begin{cases}
  \frac{1}{|\bW|}& \text{if}\;\; k_n \;\;\in \bW_{\ell_n } \\
  0& \text{Otherwise}
  \end{cases}
\end{equation}
}
The latter is  a pre-set distribution over the grid locations, uniform inside $W_{\ell_n}$. Instead of the prior $\theta_{\ell} $ the locations are generated from a top layer CG, indexed by $m$ ($\ell^{(2)}$ in the figure),
{\small
\begin{equation}
P(\ell_n | m ,\pi_{CG}) = \frac{1}{|\bW|}\cdot \sum_{\bk \in W_{m}} \pi_{CG,\bk}(\ell_n) 
\end{equation}
}
This equation also shows that the lower-levels' grid locations act as observations in the higher level. 
We use the fully factorized variational posterior $q^t( \{ k_n \}, \{ \ell_n\} , m ) = q^t( m )\cdot \prod_n \big( q^t(k_n)\cdot q^t(\ell_n) \big)$ to write the negative free energy $\mathcal{F}$ bounding the non-constant part of the loglikelihood of the data as
{\small
\begin{eqnarray}
\mathcal{F}  & = &  \sum_{t,n,k_n}  q^t(k_n)  \log \pi_{CCG,k_n}(w_n^t)  \nonumber \\  & + & \sum_{t,n,k_n,\ell_n}q^t(k_n) q^t(\ell_n) \log U_{\ell_n}^W(k_n)  \nonumber \\
& + & \sum_{t,m,\ell_n} q^t(m) q^t(\ell_n )\log \pi_{CG,m}(\ell_n) \nonumber \\
& -& \mathbb{H}\big(q(m, \{ k_n \}, \{ \ell_n \})  \big) \nonumber
\end{eqnarray}
}
We maximize $\mathcal{F}$ with the EM algorithm which iterates E- and M-steps until convergence. E:
\begin{eqnarray}
q^t(k_n = \bi)  & \propto & \big( e^{\sum_{\ell_n} q^t(\ell_n) \log U_{\ell_n}^W(\bi) }\big) \cdot \pi_{CCG,\bi}(w_n) \nonumber  \\
q^t(\ell_n = \bi)  & \propto & \big( e^{\sum_{k_n} q^t( k_n) \log U_{\bi}^W(k_n) }\big)\nonumber  \\ 
 & \cdot &  \big( e^{\sum_{m} q^t(m)\log \pi_{CG,m}(\bi)} \big) \nonumber \\
q^t(m = \bi)   & \propto & e^{\sum_n \sum_{\ell_n} q^t(\ell_n) \cdot \log h_{CG,\bi}(\ell_n)} \nonumber 
\end{eqnarray}
The M step re-estimates the model parameters using these updatedposteriors:
{\small
\begin{eqnarray}
\pi_{{\tiny CCG},\bi}(z) \hspace{-0.3cm} & \propto &\hspace{-0.1cm} \sum_t \sum_n q^t( k_n = \bi )\cdot [w_n^t = z] \nonumber \\
\pi_{{\tiny CG},\bi}(\bl) \hspace{-0.3cm} & \propto &\hspace{-0.1cm} \hat{\pi}_{{\tiny CG},\bi}(\bl) \cdot \sum_{t,n} q^t(\ell_n = \bl)\hspace{-0.cm}\cdot\hspace{-0.2cm} \sum_{\bk | \bi \in  W_\bk} \hspace{-0.2cm}\frac{ q^t( k_n = \bi )}{\hat{h}_{{\tiny CG},\bi}(\bl)}  \nonumber 
\end{eqnarray}
}
where the last (CG) update is performed analogous with \cite{cgUai}. Interestingly, training these hierarchical models stage by stage, reminiscent of deep models where such incremental learning was practically useful \cite{hintonFast}. \\
Although it has been shown that a deep neural network can be compressed into a shallow broader one through post training \cite{DBLP:journals/corr/BaC13}, the stacked ( C-)CG models can be collapsed mathematically. In this sense we can view HCG and HCCG as \emph{hierarchical learning algorithms} for CG and CCG, which are easier to visualize than deeper models. For example, for HCG in Fig. \ref{fig:gm}c-d, it is straightforward to see that the following grid defined over the original features $\{ w_n \}$,
{\small
\begin{equation}
\pi_\ell(w_n)=\sum_{\bi} \pi_{\cdot,\ell}^{(1)}(\bi)\cdot h_{CCG,\bi}^{(2)}(w_n) \label{eq:collapse}
\end{equation}
}
can be used as a single layer grid that describes the same data distribution as the two-layer model\footnote{$h_\bi$ are the grouped microtopics in the window $\bW_\bi$ - Eq. 
\ref{eq:h}}.  
However, the grids estimated from the hierarchical models should be more compact as the scattered groups of features are progressively merged in each new layer.
\emph{Learning in hierarchical models is thus more gradual and results in better local maxima, and we show below that the results are far superior to regular EM learning of the collapsed CG or CCG models.}
\vspace{-0.3cm}
\section{EXPERIMENTS}
\vspace{-0.2cm}
In all the experiments we used models with two extra layers, although, in some experiments, we found that three levels worked slightly better. In general, the optimal number of layers will depend on the particular application.

\paragraph{Likelihood comparison:} In the first experiment we compared the local maxima on models learned using the (full) MNIST data set. The two layer HCG model was first pre-trained stage-wise as, e.g., \cite{hintonFast}, by training the higher level on the posterior distribution from the lower level as the input. Then, the model was refined by further variational EM training. The procedure is repeated 20 times with different random initializations to produce twenty hierarchical models. As discussed above, these models can be collapsed to a CG model by integrating out intermediate layers (\ref{eq:collapse}). These models were then compared with twenty models learned by directly learning CG models through previously published standard EM learning algorithm starting from twenty random initializations. Despite being collapsible to the same mathematical form, the HCG models consistently produced higher likelihood than the CG models directly learned using the standard method.\emph{ In fact, each CG model created by collapsing one of the learned HCG models had log likelihood at least two standard deviations above the highest log likelihood learned by basic EM (p-value $< 10^{-20}$).} Both learning approaches used the computation time equivalent to 1000 iterations of standard EM, which was more than enough for convergence.

\paragraph{Document classification:} Next we ran test to see if the increased likelihood obtainable with a better learning algorithm translates into increased quality of representation when posterior distributions for individual text documents are considered as features in classification tasks. We considered the 20-newsgroup dataset\footnote{\url{http://www.cs.cmu.edu/afs/cs.cmu.edu/project/theo-20/www/data/news20.html}} (20N) and the Mastercook dataset\footnote{\cite{ccg}} (MC) composed by 4000 recipes divided in 15 classes. Previous work \cite{basu,sam} reduced 20-Newsgroup dataset into subsets with varying similarities and we considered the hardest subset composed by posts from the very similar newsgroups \texttt{comp.os.ms-windows}, \texttt{comp.windows.x} and \texttt{comp.graphics}. We considered the same complexities as in \cite{ccg}, using 10-fold cross validation and classified test document using maximum likelihood. Results for both datasets are shown in Tab. \ref{tab:doccl}.
{\small
\begin{table}[h!]
\centering
\begin{tabular}{l|cc|cc|c}
&CG & \textbf{HCG} & CCG & \textbf{HCCG}& linSVM\\
\hline
20N & 82,3\% & \textbf{83,5}\% & 83,4\% &\textbf{85,0}\% & 77.5\% \\
MC &38,7\% & 38,9\% & 76,2\% &\textbf{78,9}\% &  71.3\% \\
\hline
\end{tabular}
\caption{Document classification. When bold, hierarchical grids outperformed the basic grids with statistical significance (HCG p-value $=$2.01e-4, HCCG p-values $<$ 1e-3). ``linSVM'' stands for linear support vector machines which we reported as baseline.} 
\vspace{-0.4cm}
\label{tab:doccl}
\end{table}
}
\begin{figure*}[t!]
\centering
\includegraphics[width=1\textwidth]{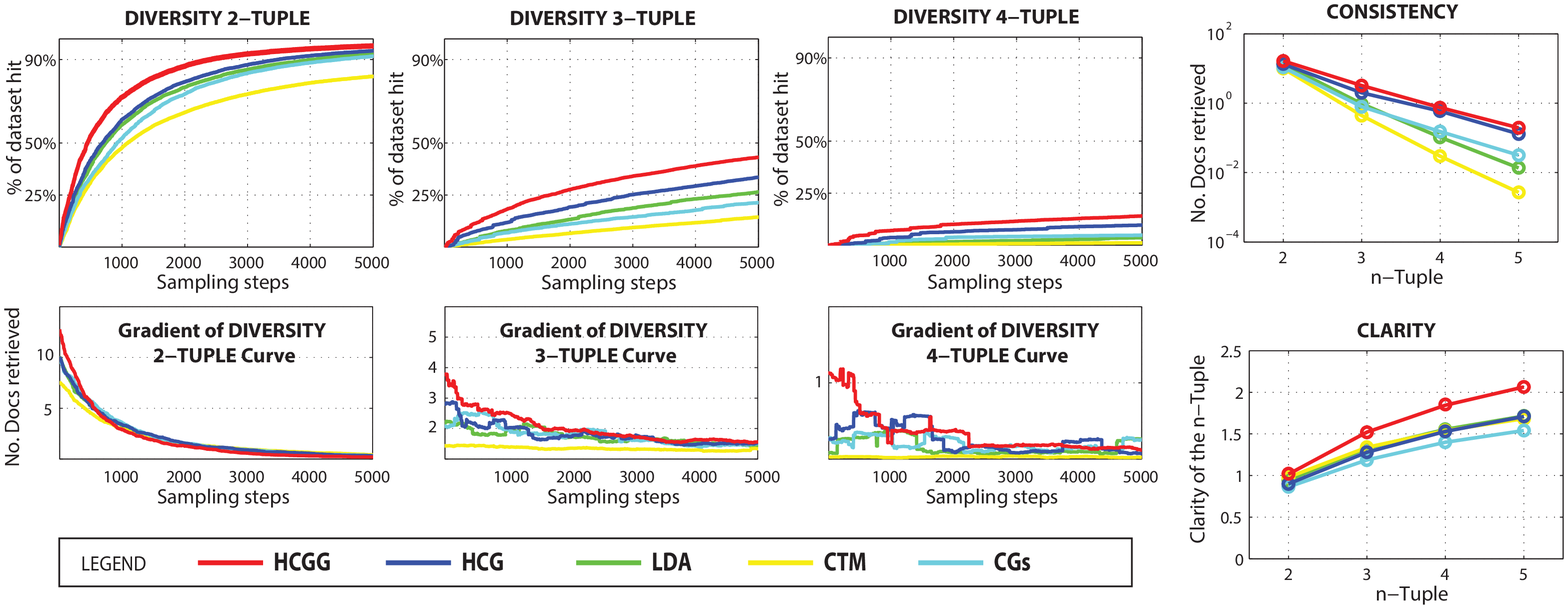}
\caption{Microtopic evaluations. We compared $32\times 32$ grids with the \emph{best} result obtained by LDA and CTM. To avoid cluttering the graph, we did not report CCG results which were found inferior to the proposed hierarchical models. We also reported the gradient of the diversity curves to show that new samples steadily continue to contribute new tuples.}
\label{fig:results}
\end{figure*}

\paragraph{Evaluation of microtopic quality using quantitative measures related to the use in visualization and indexing:}
We evaluated the coherence and the clarity of the microtopics comparing the collapsed (2 layers) hierarchical grids - HCG and HCCG with regular grids \cite{cgUai,ccg}, latent Dirichlet allocation (LDA) \cite{lda}, the correlated topic model (CTM) \cite{ctm} which allows to learn a large set of correlated topics and few non-parametric topic models \cite{paisley2012,hdp}. \\
Generative models are often evaluated in terms of perplexity. However different models, even different learning algorithms applied to the same model, are very difficult to compare \cite{asuncion2009smoothing} and better perplexity does not always indicate better quality of topics as judged by human evaluators \cite{readingTeaLeaves}. On the other hand, the subjective evaluation of topic quality is highly related to measures that have to do with data indexing, e.g. quality of word combinations when used for information retrieval. Thus we start with a novel evaluation procedure for topic models which is strongly related to information indexing and then show that we obtain similar evaluation results when we use human judgement. 
In the following experiments, we considered a corpus $\mathcal{D}$ composed of Science Magazine reports and scientific articles from the last 20 years. This is a very diverse corpus similar to the one used in \cite{ctm}. As preprocessing step, we removed stop-words and applied the Porters' stemmer algorithm \cite{porter}. We considered grids of size $16\times 16, 24\times 24, 32\times 32, 40 \times 40 $ and $48\times48$ fixing the window size to $5 \times 5$. (Previous literature showed that counting grids are only sensitive to the ratio between grid and window area, as long as windows are sufficiently big.) We varied number of topics for LDA and CTM in $\{ 10,15,\dots,100,125,150,\dots,1000 \}$. 
For each complexity we trained 5 models starting with different random initializations and we averaged the results. In each repetition, we considered a random third of this corpus, for total of roughly $|\mathcal{D}| = 12K$ documents, $Z = 20K$ different words and more than $600K$ tokens. 

To evaluate (micro)topics, we repetitively sampled \emph{k}-tuples of words and checked for consistency, diversity and clarity of the indexed content. In the following, we describe the procedure used for evaluating grids. An equivalent procedure was used to evaluate other topic models for comparison. \\
To pick a tuple $\mathcal{T}$ of $n$ words, we sampled a grid location $\hat{\ell}$. Then, we repetitively sampled the microtopic $\pi_{\hat{\ell}}$ to obtain the words in the tuple $\mathcal{T}=\{ w_1,\dots w_n \}$. We did not allow repetitions of words in the tuple. We considered $5000$ different $n=2,3,4,5$-tuples, not allowing repeated tuples. \\
Then we checked for consistency, diversity and clarity of content indexed by each tuple. 
The \textbf{consistency} is quantified in terms of the average number of documents from the dataset that contained \emph{all} words in $\mathcal{T}$. The \textbf{diversity} of indexed content is illustrated through the cumulative graph of acquired unique documents as more and more $n$-tuples are sampled and used to retrieve documents containing them. As this last curve depends on the sample order, we further repeated the process 5 times for a total of 25K different samples. Finally the \textbf{clarity} \cite{clarity}, measures the ambiguity of a query with respect to a collection of documents and it has been used to identify ineffective queries, on average, without relevance information.

Formally, the query clarity is measured as the entropy between the n-tuple and the language model $P(w)$ (unigram distributions) as $\sum_{w} P(w|T)\cdot \log_2 \frac{P(w|T)}{P(W)}$ where $P(w|T) = \sum_{d \in \mathcal{D}} P(w|\mathcal{D}) \cdot P(\mathcal{D}|T)$.
We estimated the likelihood of an individual document model generating the tuple $P(T|\mathcal{D}) = \prod_{w_t \in T}P(w_t|\mathcal{D})$ and obtain $P(\mathcal{D}|T)$ using uniform prior probabilities for documents that contains a word in the tuple, and a zero prior for the rest. Finally, to estimate $P(w|T)$ we employed MonteCarlo sampling. \\
Results are illustrated in Fig.\ref{fig:results} and should be appreciated by looking at all three measures together, as some can be over-optimized at the expense of others. The diversity curve that consistently grows as more tuples are sampled indicates that the sampled tuples belong to different subsets of the data, and are thus discriminative in segmenting the data into different clusters. The average tuple consistency, on the other hand, demonstrates that the sampled tuples do occur in large chunks of the data, demonstrating that the induced clusters are of significant size. The clarity measure shows that the clusters made of texts retrieved using different tuples have clear differentiation from the rest of the dataset in usage of all the words in the dictionary. 
We report results for the $32\times 32$ grids and the best result of LDA and CTM which peaked respectively at 80 and 60 topics. Results for other grid sizes can be found in the additional material; they are stable across complexities with slightly better performances for larger grids. \\
All grid models show good consistency of words selected as they are optimized so that documents' words map into overlapping windows. Through positioning and intersection of many related documents the words end up being arranged in a fine-grained manner so as to reflect their higher-order co-occurrence statistics. Hierarchical learning greatly improved the results despite the fact that HCCG and HCG can be reduced to (C)CGs through marginalization (\ref{eq:collapse}). \\
Overall HCCG strongly outperformed all the methods, especially with a total gain of 0.5 bits on clarity, which is around third of the score for LDA/CTM. Despite allowing for correlated topics that enable CTM to learn larger topic models, CTM trails LDA in these graphs as topics were over expanded. 
We also considered non-parametric topic models such as ``Dilan'' \cite{paisley2012} and the hierarchical Dirichlet process \cite{hdp} but their best results were poor and we did not reported them in the figure. To get an idea, both models only indexed 25\% of the content after 5000 2-Tuples samples and had a clarity lower of 0.7-1.2 bits than other topic models.

\begin{figure*}[t!]
\centering
\includegraphics[width=0.7\textwidth]{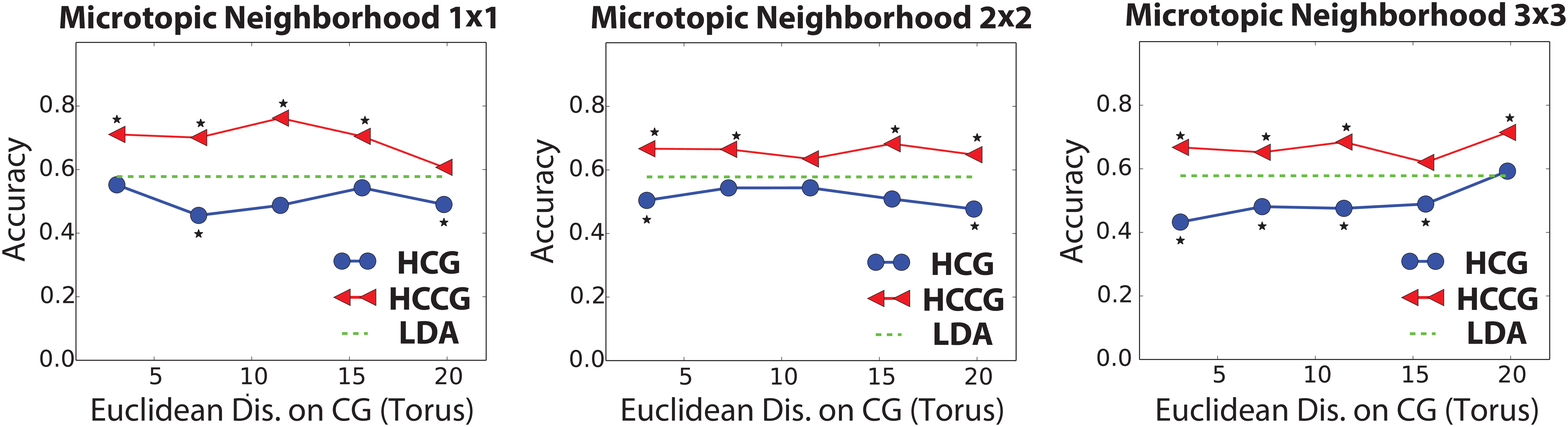}
\vspace{-0.4cm}
\caption{ Result of word intrusion task. Statistical significance is denoted by *. \emph{p-values and further details on the test are reported in the appendix}}
\vspace{-0.4cm}
\label{fig:intrusion}
\end{figure*}

\paragraph{Human judgments of topic coherence:} We next tested the quality of the inferred topics. Topic coherence is often measured based on co-occurrence of the top $k=10$ words per topic. While good as a quick sanity check of a single learned model, when this measure is used to compare models, it will favor models that lock onto top themes and distribute the rest of the words in the tails of the topic distributions. The LDA models usually have a large drop off in topic coherence when the number of topics is increased to force the model to address more correlations in the data. Indeed, using this measure, LDA topics outperform CG topics in case of small models. But as the number of topics grows, the microtopics trained by HCG significantly outperform both LDA and CG (see the appendix). A more interesting measure of topic quality, which not only depends on individual topic coherence but also on meaningful separation of different topics, requires human evaluation of \textit{word intrusions}.
In a word intrusion task \cite{readingTeaLeaves}, six randomly ordered words are presented to a human subject who then guesses which word is an outlier. In the original procedure a target topic is randomly selected and then the five words with \emph{highest} probability are picked. Then, an intruder is added to this set. It is selected at random from the low probability words of the target topic that have high probability in some other topic. Finally the six words are shuffled and presented to the subject. If the target topic shows a lack of coherence or distinction from the intruding topic, the subject will often fail to correctly identify the intruder. This task is again geared towards only getting the top words right in a topic model and ignoring the rest of the distribution, which makes it unsuitable to comparison with microtopic models which attempt to extract much more correlation from the data. Thus instead of picking the top words from each topic, we sampled the words from the target topic to create the in-group. After sampling the location of a microtopic from the grid $\hat{\ell}$, we picked three randomly chosen words from $\pi_{\hat{\ell}}$ or from the small groups of microtopics in the window of size 2$\times$2, and 3$\times$3 around $\hat{\ell}$ (The latter is equivalent to computing the window distributions $h$ using windows of smaller size than the ones used in training and should give us the indication if the granularity assumed in the window size was exaggerated: If it is then averaging of nearby topics should significantly reduce the noise due to forced topic splitting). For each of these groups we choose the intruder word using the standard procedure. If in this harder task humans can identify intruders better for microtopic models than for LDA models, this would indicate that the microtopics are not simply random subsamples of broader topics captured in $h$ and similar in entropy to LDA topics. They would be a meaningful breakup of broad topics into finer ones.
We compared LDA (known to performed better than CTM on intrusion tasks \cite{readingTeaLeaves}), HCG, and HCCG, on randomly crawled 10K Wikipedia articles and used Amazon Mechanical Turk (24000 completed tasks from 345 different people). The trained grids were of size 32 $\times$ 32 and the windows 5 $\times$ 5. The optimal LDA size was chosen using likelihood crossvalidation over the range of complexities as in the previous experiments (The peak performance there was at 80 topics).
Results are shown in Fig.\ref{fig:intrusion} as a function of the Euclidean distance on the grid of the intruder word from the topic. HCCG outperformed LDA (p-values for the 3 tasks 1.20e-11, 1.88e-5, 2.97e-05) and HCG (p-values for the 3 tasks 3.97e-18, 1.01e-11, 3.14e-19) indicating that learning microtopics is possible with a good algorithm. Overall, \emph{users were able to solve correctly 71\% of HCCG problems and only 58\% of LDA problems}. Interestingly, the performance of HCCG and HCG does not seem to depend on the distance of the intruder word: Even picking intruder word from a very close location rather than from a far away one lead to no additional confusion for the user. This shows that HCCG chops up the data into meaningful microtopics which are then combined into a large number of groups $h$ that do not over broaden the scope.
HCCG and HCG also outperformed respectively CG and CCG (see the appendix).

\begin{figure*}[t!]
\centering
\includegraphics[width=0.6\textwidth]{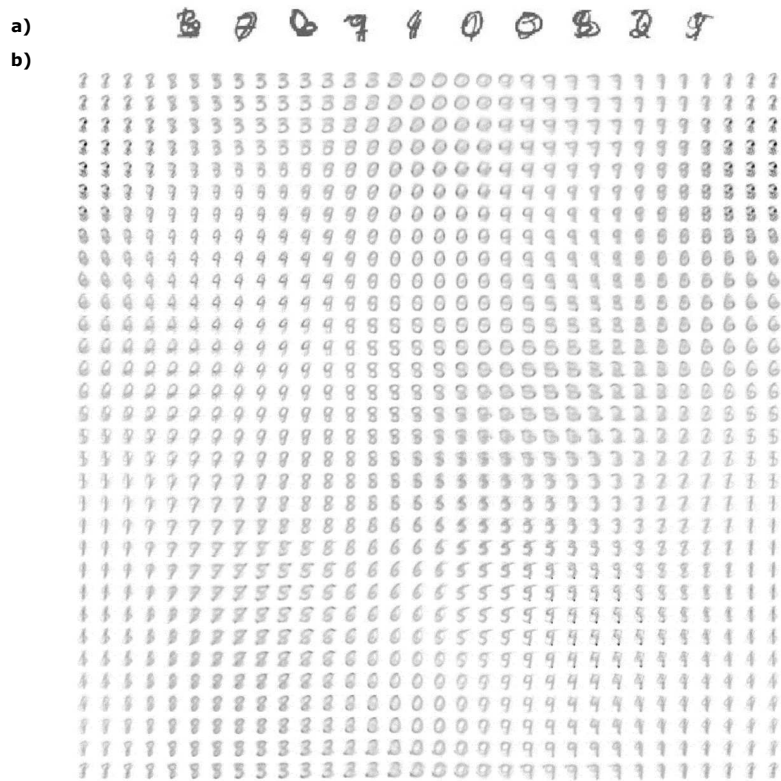}
\caption{Unsupervised learning on mixed digits}
\label{fig:digitmix}
\end{figure*}

\paragraph{Learning to separate mixed digits.} Finally, we show that an HCCG model can be used to perform a task that eludes most unsupervised \emph{and} supervised models. We created a set of 10000 $28\times 28$ images, each containing two different MNIST digits overlapped, Fig. \ref{fig:digitmix}. We trained an HCCG model consisting of five $32\times 32$ layers on this data stagewise by feeding $L^t (\ell) = \sum_{n} q^t(\ell_n=\ell)$ from one layer to the next. Windows of size 5 $\times$ 5 were used in all layers. From layer to layer, the new representations of the image consist of growing combinations of low level features $h$ from the bottom layer (sparseness of which is similar to Fig. \ref{fig:digits2}a). The hierarchical grouping is further encouraged by simply smoothing $L^t (\ell)$ with a $5\times5$ Gaussian kernel with deviation of 0.75, before feeding it to the next layer (This is motivated by the fact that nearby features in $h$ are related and so if two distant locations should be grouped, so should those locations' neighbors). Once the model is collapsed to a single HCCG grid the components no longer look like short strokes but like whole digits, mostly free of overlap: The model has learned to approximately separate the images into constitutive digits. Reasoning on overlapping digits even eludes deep neural networks trained in a supervised manner, but here we did not use the information about which two digits are present in each of the training images.
\vspace{-0.3cm}
\section{CONCLUSIONS}
\vspace{-0.2cm}
We show that with new learning algorithms based on a hierarchy of CCG models, possibly terminated on the top with a CG, it is possible to learn large grids of sparse related microtopics from relatively small datasets. These microtopics correspond to intersections of multiple documents, and are considerably narrower than what traditional topic models can achieve without overtraining on the same data. Yet, these microtopics are well formed, as both the numerical measures of consistency, diversity and clarity and the user study on 345 mechanical turkers show. Another approach to capturing sparse intersections of broader topics is through product of expert models, e.g. RBMs \cite{Salakhutdinov_replicatedsoftmax}, which consist of relatively broad topics but model the data through intersections rather than admixing. RBMs are also often stacked into deep structures. In future work it would be interesting to compare these models, though the tasks we used here would have to be somewhat changed to focus on the intersection modeling, rather than the topic coherence (as this is not what RBM topics are optimized for). HCCG and HCG models have a clear advantage in that it is easy to visualize how the data is represented, which is useful both to end users in HCI applications, and to machine learning experts during model development and debugging. Another parallel between the stacks of CCGs and other deep models is that the uniform connectivity of units is directly enforced through window constraints, rather than encouraged by dropout. Finally, in this specific context we illustrate a broader phenomenon that requires more methodical and broader treatment by the machine learning community. A more complex (deeper) model showed here large advantages in terms of training likelihood, but these advantages were \emph{not} due to the expanded parameter space, because the resulting model is equivalent to a collapsed single layer model. Rather than being a reflection of increased representational abilities of the model, better likelihoods were thus the result of better fitting algorithm that consists of training a deep model (and then collapsing it into a simpler but equivalent parameterization). Similar phenomena are likely regularly encountered elsewhere in machine learning, but not always recognized as such, as in the absence of the full knowledge of the extrema of the fitting criterion, an increase in performance is often inappropriately ascribed to better modeling rather than better model fitting.


\clearpage
{\small
\begin{thebibliography}{10}

\bibitem{cgUai}
Jojic, N., Perina, A.:
\newblock Multidimensional counting grids: Inferring word order from disordered
  bags of words.
\newblock In: Proceedings of conference on Uncertainty in artificial
  intelligence (UAI). (2011)  547--556

\bibitem{ccg}
Perina, A., Jojic, N., Bicego, M., Truski, A.:
\newblock Documents as multiple overlapping windows into grids of counts.
\newblock In Burges, C., Bottou, L., Welling, M., Ghahramani, Z., Weinberger,
  K., eds.: Advances in Neural Information Processing Systems 26.
\newblock Curran Associates, Inc. (2013)  10--18

\bibitem{perinaKDD}
Perina, A., Kim, D., Truski, A., Jojic, N.:
\newblock Skim-reading thousands of documents in one minute: Data indexing and
  visualization for multifarious search.
\newblock In: Workshop on Interactive Data Exploration and Analytics (IDEA'14)
  at KDD. (2014)

\bibitem{kusner-etal}
Matt J.~Kusner, Yu~Sun, N.I.K.K.Q.W.:
\newblock From word embeddings to document distances.
\newblock In: Proceedings of the International Conference on Machine Learning.
  (2015)

\bibitem{ssvi}
Hoffman, M., Blei, D.:
\newblock Structured stochastic variational inference.
\newblock CoRR \textbf{abs/1404.4114} (2014)

\bibitem{topic_models_practice}
Yi, X., Allan, J.:
\newblock A comparative study of utilizing topic models for information
  retrieval.
\newblock In: Proceedings of the 31th European Conference on IR Research on
  Advances in Information Retrieval. ECIR '09, Berlin, Heidelberg,
  Springer-Verlag (2009)  29--41

\bibitem{lda}
Blei, D., Ng, A., Jordan, M.:
\newblock Latent dirichlet allocation.
\newblock Journal of machine Learning Research \textbf{3} (2003)  993--1022

\bibitem{ctm}
Blei, D.M., Lafferty, J.D.:
\newblock Correlated topic models.
\newblock In: NIPS. (2005)

\bibitem{Crow}
Crow, F.C.:
\newblock Summed-area tables for texture mapping.
\newblock In: Proceedings of the 11th Annual Conference on Computer Graphics
  and Interactive Techniques. SIGGRAPH '84, New York, NY, USA, ACM (1984)
  207--212

\bibitem{hintonFast}
Hinton, G., Osinero, S.:
\newblock A fast learning algorithm for deep belief nets.
\newblock Neural Computation \textbf{18} (2006)

\bibitem{DBLP:journals/corr/BaC13}
Ba, L.J., Caurana, R.:
\newblock Do deep nets really need to be deep?
\newblock CoRR \textbf{abs/1312.6184} (2013)

\bibitem{basu}
Banerjee, A., Basu, S.:
\newblock Topic models over text streams: a study of batch and online
  unsupervised learning.
\newblock In: In Proc. 7th SIAM Int’l. Conf. on Data Mining. (2007)

\bibitem{sam}
Reisinger, J., Waters, A., Silverthorn, B., Mooney, R.J.:
\newblock Spherical topic models.
\newblock In: ICML '10: Proceedings of the 27th international conference on
  Machine learning. (2010)

\bibitem{paisley2012}
Paisley, J., Wang, C., Blei, D.M.:
\newblock The discrete infinite logistic normal distribution.
\newblock Bayesian Analysis \textbf{7} (2012)  997--1034

\bibitem{hdp}
Teh, Y.W., Jordan, M.I., Beal, M.J., Blei, D.M.:
\newblock Hierarchical dirichlet processes.
\newblock Journal of the American Statistical Association \textbf{101} (2004)

\bibitem{asuncion2009smoothing}
Asuncion, A., Welling, M., Smyth, P., Teh, Y.W.:
\newblock On smoothing and inference for topic models.
\newblock In: In Proceedings of Uncertainty in Artificial Intelligence. (2009)

\bibitem{readingTeaLeaves}
Chang, J., Boyd-Graber, J.L., Gerrish, S., Wang, C., Blei, D.M.:
\newblock Reading tea leaves: How humans interpret topic models.
\newblock In: NIPS. (2009)

\bibitem{porter}
Porter, M.F.:
\newblock Readings in information retrieval.
\newblock Morgan Kaufmann Publishers Inc., San Francisco, CA, USA (1997)
  313--316

\bibitem{clarity}
Cronen-Townsend, S., Croft, W.B.:
\newblock Quantifying query ambiguity.
\newblock In: Proceedings of the Second International Conference on Human
  Language Technology Research. HLT '02, San Francisco, CA, USA, Morgan
  Kaufmann Publishers Inc. (2002)  104--109

\bibitem{Salakhutdinov_replicatedsoftmax}
Salakhutdinov, R., Hinton, G.:
\newblock Replicated softmax: an undirected topic model.
\newblock In: In Advances in Neural Information Processing Systems, NIPS.
  (2009)

\end{thebibliography}
}
\end{document}